\documentclass[11pt,a4paper]{article}

\usepackage[acceptedWithA]{tacl2018v2}
\usepackage{times,latexsym}
\usepackage{url}
\usepackage[T1]{fontenc}

\usepackage{amsmath}
\usepackage{multirow}
\usepackage{graphicx}
\graphicspath{ {figure/} }

\usepackage{bm}
\usepackage{verbatim}

\title{Integrating Weakly Supervised Word Sense Disambiguation\\into Neural Machine Translation}

\author{Xiao Pu\Thanks{Work conducted while at the
                      Idiap Research Institute.}\\
  Nuance Communications\\
  J\"{u}licher Str.\ 376\\
	52070 Aachen, Germany \\
  {\sf xiao.pu@nuance.com} \\\And
	Nikolaos Pappas\\
  Idiap Research Institute\\
  Rue Marconi 19, CP 592\\
  1920 Martigny, Switzerland\\
  {\sf nikolaos.pappas@idiap.ch} \\\AND
	James Henderson\\
  Idiap Research Institute\\
	Rue Marconi 19, CP 592\\
  1920 Martigny, Switzerland\\
  {\sf james.henderson@idiap.ch} \And
	Andrei Popescu-Belis \\
  HEIG-VD / HES-SO\\
  Route de Cheseaux 1, CP 521\\
  1401 Yverdon-les-Bains, Switzerland\\
  {\sf andrei.popescu-belis@heig-vd.ch} }

\date{} 

\begin{document}
\maketitle
\begin{abstract}
This paper demonstrates that word sense disambiguation (WSD) can improve neural machine translation (NMT) by widening the source context considered when modeling the senses of potentially ambiguous words.  We first introduce three adaptive clustering algorithms for WSD, based on $k$-means, Chinese restaurant processes, and random walks, which are then applied to large word contexts represented in a low-rank space and evaluated on SemEval shared-task data.  We then learn word vectors jointly with sense vectors defined by our best WSD method, within a state-of-the-art NMT system.  We show that the concatenation of these vectors, and the use of a sense selection mechanism based on the weighted average of sense vectors, outperforms several baselines including sense-aware ones.  This is demonstrated by translation on five language pairs.  The improvements are above one BLEU point over strong NMT baselines, +4\% accuracy over all ambiguous nouns and verbs, or +20\% when scored manually over several challenging words.
\end{abstract}

\section{Introduction}
\label{intro}

The correct translation of polysemous words remains a challenge for machine translation (MT).  While some translation options may be interchangeable, substantially different senses of source words must generally be rendered by different words in the target language.  Hence, an MT system should identify -- implicitly or explicitly -- the correct sense conveyed by each occurrence in order to generate an appropriate translation.  For instance, in the following sentence from Europarl, the translation of `deal' should convey the sense `to handle' (in French `\emph{traiter}') and not `to cope' (in French `\emph{rem\'{e}dier}', which is wrong):

\begin{description} \setlength{\itemsep}{0pt}
\item[\textbf{Source:}] How can we guarantee the system of prior notification for high-risk products at ports that have the necessary facilities to \emph{deal} with them?
\item[\textbf{Reference translation:}] Comment pouvons-nous garantir le syst\`{e}me de notification pr\'{e}alable pour les produits pr\'{e}sentant un risque \'{e}lev\'{e} dans les ports qui disposent des installations n\'{e}cessaires pour \emph{traiter} ces produits ?
\item[\textbf{Baseline NMT:}] $[\ldots]$ les ports  qui disposent des moyens n\'{e}cessaires pour y \emph{rem\'{e}dier}~?
\item[\textbf{Sense-aware NMT:}] $[\ldots]$ les ports qui disposent des installations n\'{e}cessaires pour les \emph{traiter}~? 
\end{description}

\noindent
Current MT systems perform word sense disambiguation implicitly, based on co-occurring words in a rather limited context.  In phrase-based statistical MT, the context size is related to the order of the language model (often between 3 and 5) and to the length of n-grams in the phrase table (seldom above~5).  In attention-based neural MT (NMT), the context extends to the entire sentence, but multiple word senses are not modeled explicitly.  The implicit sense information captured by word representations used in NMT leads to a bias in the attention mechanism towards dominant senses. Therefore, the NMT decoders cannot clearly identify the contexts in which one word sense should be used instead of another one. Hence, while NMT can use local constraints to translate `great rock band' into French as `\emph{superbe groupe de rock}' rather than `\emph{grande bande de pierre}' -- thus correctly assigning the musical rather than geological sense to `rock' -- it fails to do so for word senses which require larger contexts. 
 
In this paper, we demonstrate that the explicit modeling of word senses can be helpful to NMT by using combined vector representations of word types and senses, which are inferred from contexts that are larger than that of state-of-the-art NMT systems. We make the following contributions: 
\begin{itemize}
\item Weakly supervised word sense disambiguation (WSD) approaches integrated into NMT, based on three adaptive clustering methods and operating on large word contexts.
\item Three sense selection mechanisms for integrating WSD into NMT, respectively based on top, average, and weighted average (i.e., attention) of word senses.
\item Consistent improvements against baseline NMT on five language pairs: from English into Chinese, Dutch, French, German, and Spanish.
\end{itemize}

The paper is organized as follows.  In Section~\ref{section:wsd}, we present three adaptive WSD methods based on $k$-means clustering, the Chinese restaurant process, and random walks.  In Section~\ref{sec:nmt_inter}, we present three sense selection mechanisms which integrate the word senses into NMT.  The experimental details appear in Section~\ref{sec:data}, while the results concerning the optimal parameter settings are presented in Section~\ref{sec:optimal-parameters}, where we also show that our WSD component is competitive on the SemEval 2010 shared task.  Section~\ref{sec:mt-results} presents our results: the BLEU scores increase by about one point with respect to a strong NMT baseline, the accuracy of ambiguous noun and verb translation improves by about 4\%, while a manual evaluation of several challenging and frequent words shows an improvement of about 20\%.  A discussion of related work appears finally in Section~\ref{sec:related-work}.

\section{Adaptive Sense Clustering for MT}
\label{section:wsd}

In this section, we present the three unsupervised or weakly supervised WSD methods used in our experiments, which aim at clustering different occurrences of the same word type according to their senses.  We first consider all nouns and verbs in the source texts that have more than one sense in WordNet, and extract from there the definition of each sense and, if available, the example.  For each occurrence of such nouns or verbs in the training data, we use word2vec to build word vectors for their contexts, i.e.\ neighboring words.  All vectors are passed to an unsupervised clustering algorithm, possibly instantiated with WordNet definitions or examples.  The resulting clusters can be numbered and used as labels, or their centroid word vector can be used as well, as explained in Section~\ref{sec:nmt_inter}.

This approach answers several limitations of previous supervised or unsupervised WSD methods.  On the one hand, supervised methods require data with manually sense-annotated labels and are thus limited to typically small subsets of all word types, e.g.\ up to a hundred of content words targeted in SemEval 2010\footnote{\url{www.cs.york.ac.uk/semeval2010_WSI}} \cite{manandhar2010semeval} and up to a thousand words in SemEval 2015 \cite{moro2015}.  In contrast, our method does not require labeled texts for training, and applies to all word types with multiple senses in WordNet (e.g.\ nearly four thousands for some data sets, see Table~\ref{mt_data}).  On the other hand, unsupervised methods often pre-define the number of possible senses for all ambiguous words before clustering their occurrences, and do not adapt to what is actually observed in the data; as a result, the senses are often too fine-grained for the needs of MT, especially for a particular domain. In contrast, our model learns the number of senses for each analyzed ambiguous word directly from the data.

\subsection{Definitions and Notations}

For each noun or verb type $W_{t}$ appearing in the training data, as identified by the Stanford POS tagger,\footnote{\url{nlp.stanford.edu/software}} we extract the senses associated to it in WordNet\footnote{\url{wordnet.princeton.edu/}} \cite{wordnet1998book} using NLTK.\footnote{\url{www.nltk.org/howto/wordnet.html}}  Specifically, we extract the set of definitions $D_t = \{d_{tj} | j=1,\ldots,m_{t}\}$ and the set of examples of use $E_t = \{e_{tj} | j=1,\ldots,n_{t}\}$, each of them containing multiple words.  While most of the senses are accompanied by a definition, only about half of them also include an example of use.

Definitions $d_{tj}$ and examples $e_{tj}$ are represented by vectors defined as the average of the word embeddings over all the words constituting them (except stopwords).  Formally, these vectors are $\mathbf{d}_{tj} = (\sum_{w_{l}\in d_{tj}} \mathbf{w}_{l})/ |d_{tj}|$ and $\mathbf{e}_{tj} = (\sum_{w_{l}\in e'_{tj}} \mathbf{w}_{l}) / |e'_{tj}|$, respectively, where $|d_{tj}|$ is the number of tokens of the definition.  While the entire definition $d_{tj}$ is used to build the $\mathbf{d}_{tj}$ vector, we do not consider all words in the example $e_{tj}$, but limit the sum to a fragment $e'_{tj}$ contained in a window of size $c$ centered around the considered word, to avoid noise from long examples.  Hence, we divide by the number of words in this window, noted $|e'_{tj}|$.   All the word vectors $\mathbf{w}_{l}$ above are pre-trained word2vec embeddings from Google\footnote{\url{code.google.com/archive/p/word2vec/}} \cite{mikolov2013distributed}.  If $\mathit{dim}$ is the dimensionality of the word vector space, then all vectors $\mathbf{w}_{l}$, $\mathbf{d}_{tj}$, and $\mathbf{e}_{tj}$ are in~$\mathcal{R}^\mathit{dim}$.  Each definition vector $\mathbf{d}_{tj}$ or example vector $\mathbf{e}_{tj}$ for a word type $W_t$ is considered as a center vector for each sense during the clustering procedure.

Turning now to tokens, each word occurrence $w_{i}$ in a source sentence is represented by the average vector $\mathbf{u}_{i}$ of the words from its context, i.e.\ a window of $c$ words centered on $w_{i}$, $c$ being an even number.  We calculate the vector $\mathbf{u}_{i}$ for $w_{i}$ by averaging vectors from $c/2$ words before $w_i$ and from $c/2$ words after it.  We stop nevertheless at the sentence boundaries, and filter out stopwords before averaging.

\subsection{Clustering Word Occurrences by Sense}
\label{sec:clustering}

We adapt three clustering algorithms to our needs for WSD applied to NMT.  The objective is to cluster all occurrences $w_i$ of a given word type $W_t$, represented as word vectors $\mathbf{u}_{i}$, according to the similarity of their senses, as inferred from the similarity of the context vectors.  We compare the algorithms empirically in Section~\ref{sec:optimal-parameters}.

\textbf{$k$-means Clustering.} The original $k$-means algorithm \cite{kmeans} aims to partition a set of items, which are here tokens $w_{1}, w_{2}, \dots, w_{n}$ of the same word type $W_t$, represented through their embeddings $\mathbf{u}_{1}, \mathbf{u}_{2}, \dots, \mathbf{u}_{n}$ where $\mathbf{u}_{i} \in \mathcal{R}^\mathit{dim}$.  The goal of $k$-means is to partition (or cluster) these vectors into $k$ sets $S=\{S_{1}, S_{2}, \ldots, S_{k}\}$ so as to minimize the within-cluster sum of squared distances to each centroid $\bm{\mu}_{i}$:
\begin{equation}
S = \arg\min_{S}\sum_{i=1}^{k}\sum_{\mathbf{u} \in S_{i}} ||\mathbf{u}-\bm{\mu}_{i}||^{2} .
\label{eq:kmeans}
\end{equation}
At the first iteration, when there are no clusters yet, the algorithm selects $k$ random points as centroids of the $k$ clusters. Then, at each subsequent iteration $t$, the algorithm calculates for each candidate cluster a new centroid of the observations, defined as their average vector, as follows:
\begin{equation}
\bm{\mu}_{i}^{\,t+1} = \frac{1}{|S_{i}^{t}|} \sum_{\mathbf{u}_{j}\in S_{i}^{t}} \mathbf{u}_{j} .
\end{equation}

In an earlier application of $k$-means to phrase-based statistical MT, but not neural MT, we made several modifications to the original $k$-means algorithm, to make it adaptive to the word senses observed in training data \citep{pu-pappas-popescubelis:2017:WMT}.  We maintain these changes and summarize them briefly here.  The initial number of clusters $k_t$ for each ambiguous word type $W_t$ is set to the number of its senses in WordNet, either considering only the senses that have a definition or those that have an example.  The centroids of the clusters are initialized to the vectors representing the senses from WordNet, either using their definition vectors $\mathbf{d}_{tj}$ or their example vectors $\mathbf{e}_{tj}$.  These initializations are thus a form of weak supervision of the clustering process.   

Finally, and most importantly, after running the $k$-means algorithm, the number of clusters for each word type is reduced by removing the clusters that contain fewer than 10 tokens and assigning their tokens to the closest large cluster. `Closest' is defined in terms of the cosine distance between $\mathbf{u}_i$ and their centroids.  The final number of clusters thus depends on the observed occurrences in the training data (which is the same data as for MT), and avoids modeling infrequent senses which are difficult to translate anyway.  When used in NMT, in order to assign each new token from the test data to a cluster, i.e.\ to perform WSD, we select the closest centroid, again in terms of cosine distance.

\textbf{Chinese Restaurant Process.}
The Chinese Restaurant Process (CRP) is an unsupervised method considered as a practical interpretation of a Dirichlet process \cite{ferguson1973bayesian} for non-parametric clustering. In the original analogy, each token is compared to a customer in a restaurant, and each cluster is a table where customers can be seated.  A new customer can choose to sit at a table with other customers, with a probability proportional to the numbers of customers at that table, or sit at a new, empty table.  In an application to multi-sense word embeddings, Li and Jurafsky \shortcite{li-jurafsky:2015:EMNLP} proposed that the probability to ``sit at a table'' should also depend on the contextual similarity between the token and the sense modeled by the table.  We build upon this idea and bring several modifications that allow for an instantiation with sense-related knowledge from WordNet, as follows.

For each word type $W_{t}$ appearing in the data, we start by fixing the maximal number $k_t$ of senses or clusters as the number of senses of $W_t$ in WordNet.  This avoids an unbounded number of clusters (as in the original CRP algorithm) and the risk of cluster sparsity by setting a non-arbitrary limit based on linguistic knowledge.  Moreover, we define the initial centroid of each cluster as the word vector corresponding either to the definition $\mathbf{d}_{tj}$ of the respective sense, or alternatively to the example $\mathbf{e}_{tj}$ illustrating the sense.

For each token $w_{i}$ and its context vector $\mathbf{u}_i$ the algorithm decides whether the token is assigned to one of the sense clusters $S_{j}$ to which previous tokens have been assigned, or whether it is assigned to a new empty cluster, by selecting the option which has the highest probability, which is computed as follows:
\begin{equation}\label{eq:whole}
P\propto%
\begin{cases}
N_{j} ( \lambda_{1}  s(\mathbf{u}_{i},\mathbf{d}_{tj}) + \lambda_{2}  s(\mathbf{u}_{i},\bm{\mu}_{j}) )
\\ \hspace{5mm}\text{if $N_{j} \ne 0$ (non-empty sense)}\\
\gamma s(\mathbf{u}_{i}, \mathbf{d}_{tj}) \\ \hspace{5mm}\text{if $N_{j} = 0$ (empty sense).}
\end{cases}
\end{equation}

\noindent
In other words, for a non-empty sense, the probability is proportional to the popularity of the sense (number of tokens it already contains, $N_j$) and to the weighted sum of two cosine similarities $s(\cdot,\cdot)$: one between the context vector $\mathbf{u}_i$ of the token and the definition of the sense $\mathbf{d}_{tj}$, and another one between $\mathbf{u}_i$ and the average context vector of the tokens already assigned to the sense ($\bm{\mu}_{j}$).  These terms are weighted by the two hyper-parameters $\lambda_{1}$ and $\lambda_{2}$.  For an empty sense, only the second term is used, weighted by the $\gamma$ hyper-parameter.

\textbf{Random Walks.}
Finally, we also consider for comparison a WSD method based on random walks on the WordNet knowledge graph \citep{agirre2009personalizing,Agirre:2014:CL} available from the UKB toolkit.\footnote{\url{ixa2.si.ehu.es/ukb}.  Strictly speaking, this is the only genuine WSD method, as the two previous ones pertain to sense induction rather than disambiguation.  However, for simplicity, we will refer to all of them as WSD.}  In the graph, senses correspond to nodes and the relationships or dependencies between pairs of senses correspond to the edges between those nodes.  From each input sentence, we extract its content words (nouns, verbs, adjectives, and adverbs) that have an entry in the WordNet weighted graph.  The method calculates the probability of a random walk over the graph from a target word's sense ending on any other sense in the graph, and determines the sense with the highest probability for each analyzed word.  In this case, the random walk algorithm is PageRank \citep{grin1998anatomy}, which computes a relative structural importance or `rank' for each node.

\section{Integration with Neural MT}
\label{sec:nmt_inter}

\subsection{Baseline Neural MT model} 
We now present several models integrating WSD into NMT, starting from an attention-based NMT baseline \cite{bahdanau2014neural,luong-pham-manning:2015:EMNLP}.  Given a source sentence $X$ with words $w^x$, $X = (w_{1}^{x},w_{2}^{x},...,w_{T}^{x})$, the model computes a conditional distribution over translations, expressed as $p(Y = (w_{1}^{y},w_{2}^{y},...,w_{T'}^{y})|X)$.  The neural network model consists of an encoder, a decoder and an attention mechanism.  First, each source word $w_{t}^{x} \in V$ is projected from a one-hot word vector into a continuous vector space representation $\mathbf{x}_t$.  Then, the resulting sequence of word vectors is read by the bidirectional encoder which consists of forward and backward recurrent networks (RNNs). The forward RNN reads the sequence in left-to-right order, i.e.\
$\overrightarrow{\mathbf{h}}_{t} = \overrightarrow{\phi}(\overrightarrow{\mathbf{h}}_{t-1}, \mathbf{x_{t}})$,
while the backward RNN reads it right-to-left:
$\overleftarrow{\mathbf{h}}_{t} = \overleftarrow{\phi}(\overleftarrow{\mathbf{h}}_{t+1}, \mathbf{x_{t}})$.

The hidden states from the forward and backward RNNs are concatenated at each time step $t$ to form an `annotation' vector $\mathbf{h_{t}} = [\overrightarrow{\mathbf{h_{t}}};\overleftarrow{\mathbf{h_{t}}}]$. Taken over several time steps, these vectors form the `context' i.e.\ a tuple of annotation vectors $C = (\mathbf{h_{1}},\mathbf{h_{2}},...,\mathbf{h_{T}})$. The recurrent activation functions $\overrightarrow{\phi}$ and $\overleftarrow{\phi}$ are either long short-term memory units (LSTM) or gated recurrent units (GRU).

The decoder RNN maintains an internal hidden state $z_{t'}$.  After each time step $t'$, it first uses the attention mechanism to weight the annotation vectors in the context tuple $C$. The attention mechanism takes as input the previous hidden state of the decoder and one of the annotation vectors, and returns a relevance score $e_{t',t} = f_\mathrm{ATT}(\mathbf{z_{t'-1}},\mathbf{h_{t}})$.  These scores are normalized to obtain attention scores:
\begin{equation}
\alpha_{t',t} = exp(e_{t',t}) / \sum_{k=1}^{T} exp(e_{t',k}).
\end{equation}
These scores serve to compute a weighted sum of annotation vectors $\mathbf{c_{t'}} = \sum_{t=1}^{T}\alpha_{t',t}\mathbf{h}_{t}$, which are used by the decoder to update its hidden state:
\begin{equation}
z_{t'} = \phi_{z}(z_{t'-1},\mathbf{y}_{t'-1},\mathbf{c}_{t'}).
\end{equation}
Similarly to the encoder, $\phi_{z}$ is implemented as either an LSTM or GRU and $\mathbf{y}_{t'-1}$ is the target-side word embedding vector corresponding to word~$w^y$.

\subsection{Sense-aware Neural MT Models}

To model word senses for NMT, we concatenate the embedding of each token with a vector representation of its sense, either obtained from one of the clustering methods presented in Section~\ref{section:wsd}, or learned during encoding, as we will explain.  In other words, the new vector $\mathbf{w'}_{i}$ representing each source token $w_{i}$ consists of two parts: $\mathbf{w'}_{i} = [\mathbf{w}_{i}\ ;\ \ \bm{\mu}_{i}] \label{eq:concat}$, where $\mathbf{w}_{i}$ is the word embedding learned by the NMT, and $\bm{\mu}_{i}$ is the sense embedding obtained from WSD or learned by the NMT.  To represent these senses, we create two dictionaries, one for words and the other one for sense labels, which will be embedded in a low-dimensional space, before the encoder.  We propose several models for using and/or generating sense embeddings for NMT, named and defined as follows. 

\textbf{Top sense (\emph{TOP}).}
In this model, we directly use the sense selected for each token by one of the WSD systems above, and use the embeddings of the respective sense as generated by NMT after training. 

\textbf{Weighted average of senses (\emph{AVG)}.}
Instead of fully trusting the decision of a WSD system (even one adapted to MT), we consider all listed senses and the respective cluster centroids learned by the WSD system. Then we convert the distances $d_{l}$ between the input token vector and the centroid of each sense $S_l$ 
into a normalized weight distribution either by a linear or a logistic normalization:
\begin{equation}\label{eq_logistic}
\omega_{j} = \frac{1-d_{j}}{\sum_{1 \leq l \leq k} d_{l}}\ \mathrm{or}\ \ 
\omega_{j} = \frac{e^{-d_{j}^{2}}}{\sum_{1 \leq l \leq k} e^{-d_{l}^2}}\ \mathrm{,}
\end{equation}
where $k$ is the total number of senses of token $w_{i}$.  The sense embedding for each token is computed as the weighted average of all sense embeddings:
\begin{equation}
\bm{\mu}_{i} = \sum_{1 \leq j \leq k} \omega_{j} \bm{\mu}_{ij}.
\label{eq:average_weight}
\end{equation}

\textbf{Attention-based sense weights (\emph{ATT}).} Instead of obtaining the weight distribution from the centroids computed by WSD, we also propose to dynamically compute the probability of relatedness to each sense based on the current word and sense embeddings during encoding, as follows.  Given a token $w_{i}$, we consider all the other tokens in the sentence $(w_{1},\ldots,w_{i-1}, w_{i+1},\ldots,w_{L})$ as the context of $w_{i}$, where $L$ is the length of the sentence. We define the context vector of $w_{i}$ as the mean of all the embeddings $\bm{u}_j$ of the words $w_j$, i.e.\ $\mathbf{u}_{i} = (\sum_{l \neq i}\bm{u}_{l}) / (L-1)$.   Then, we compute the similarity $f(\mathbf{u}_{i}, \bm{\mu}_{ij})$ between each sense embedding $\bm{\mu}_{ij}$ and the context vector $\mathbf{u}_{i}$ using an additional attention layer in the network, with two possibilities which will be compared empirically: 
\begin{equation} 
f(\mathbf{u}_{i}, \bm{\mu}_{ij}) =  \upsilon^{T} tanh(W\mathbf{u}_{i} + U\bm{\mu}_{ij})\\
\label{eq:attention}
\end{equation}
\begin{equation}
\mathrm{or}\ f(\mathbf{u}_{i}, \bm{\mu}_{ij}) =  \mathbf{u}_{i}^{T}W\bm{\mu}_{ij}.
\end{equation}
The weights $\omega_j$ are now obtained through the following softmax normalization:
\begin{equation}
\omega_{j} = \frac{e^{f(\mathbf{u}_{i}, \bm{\mu}_{ij})}}{\sum_{1 \leq l \leq k} e^{f(\mathbf{u}_{i}, \bm{\mu}_{il})}}.
\label{eq:softmax}
\end{equation}
Finally, the average sense embedding is obtained as in Eq.~\eqref{eq:average_weight} above, and is concatenated to the word vector $\mathbf{u}_{i}$.

\textbf{\emph{ATT} model with initialization of embeddings (\emph{ATT}$\bm{_\mathit{ini}}$).}
The fourth model is similar to the \emph{ATT} model, with the difference that we initialize the embeddings of the source word dictionary using the word2vec vectors of the word types, and the embeddings of the sense dictionary using the centroid vectors obtained from $k$-means.

\begin{table}[ht]
\setlength{\tabcolsep}{3pt}
\centering
\begin{tabular}{|c|c|c|c|c|c|c|}
\hline
\multirow{2}{*}{TL}      & \multirow{2}{*}{Train} & \multirow{2}{*}{Dev} & \multirow{2}{*}{Test} & \multicolumn{2}{c|}{Labels} & \multirow{2}{*}{Words} \\ \cline{5-6}
                       &                           &                              &                          & Nouns        & Verbs        &                              \\ \hline
\multirow{2}{*}{FR} & 0.5M                      & 5k                           & 50k                      & 3,910        & 1,627        & 2,006                        \\ \cline{2-7} 
                       & 5.3M                      & 4,576                         & 6,003                     & 8,276        & 3,059        & 3,876                        \\ \hline
\multirow{2}{*}{DE} & 0.5M                      & 5k                           & 50k                      & 3,885        & 1,576        & 1,976                        \\ \cline{2-7} 
                       & 4.5M                         & 3,000                            & 5,172                        & 7,520        & 1,634        & 3,194                        \\ \hline
\multirow{2}{*}{ES} & 0.5M                      & 5k                           & 50k                      & 3,862        & 1,627        & 1,987                        \\ \cline{2-7} 
                       & 3.9M                      & 4,576                         & 6,003                     & 7,549        & 2,798        & 3,558                        \\ \hline
ZH   & 0.5M                      & 5K                         & 50K                     & 3,844        & 1,475        & 1,915                        \\ \hline
NL  & 0.5M                      & 5K                         & 50K                     & 3,915        & 1,647        & 2,210                        \\ \hline                                              
\end{tabular}
\caption{Size of data sets used for machine translation from English to five different target languages (TL).} 
\label{mt_data}
\end{table}
 
\section{Data, Metrics and Implementation}
\label{sec:data}

\textbf{Datasets.} We train and test our sense-aware MT systems on the data shown in Table~\ref{mt_data}: the UN Corpus\footnote{\url{www.uncorpora.org}} \cite{rafalovitch2009united} and the Europarl Corpus\footnote{\url{www.statmt.org/europarl}} \cite{koehn2005europarl}. We first experiment with our models using the same dataset and protocol as in our previous work \citep{pu-pappas-popescubelis:2017:WMT}, to enable comparisons with phrase-based statistical MT systems, for which the sense of each ambiguous source word was modeled as a factor.  Moreover, in order to make a better comparison with other related approaches, we train and test our sense-aware NMT models on large datasets from WMT shared tasks over three language pairs (EN/DE, EN/ES and EN/FR).

The dataset used in our previous work consists of 500k parallel sentences for each language pair, 5k for development and 50k for testing. 
The data originates from UN for EN/ZH, and from Europarl for the other pairs. The source sides of these sets contain around 2,000 different English word forms (after lemmatization) that have more than one sense in WordNet.  Our WSD system generates ca.\ 3.8k different noun labels and 1.5k verb labels for these word forms. 

The WMT datasets additionally used in this paper are the following ones.  First, we use the complete EN/DE set from WMT 2016 \cite{bojar-EtAl:2016:WMT1} with a total of ca.\ 4.5M sentence pairs.  In this case, the development set is NewsTest 2013, and the testing set is made of NewsTest 2014 and 2015.  Second, for EN/FR and EN/ES, we use data from WMT 2014 \cite{bojar-EtAl:2014:W14-33}\footnote{We selected the data from different years of WMT because the EN/FR and EN/ES pairs were only available in WMT 2014.} with 5.3M sentences for EN/FR and 3.8M sentences for EN/ES. Here, the development sets are NewsTest 2008 and 2009, while the testing sets are NewsTest 2012 and 2013 for both language pairs. The source sides of these larger additional sets contain around 3,500 unique English word forms with more than one sense in WordNet, and our system generates ca.\ 8k different noun labels and 2.5k verb labels for each set. 

Finally, for comparison purposes and model selection, we use the WIT$^\mathrm{3}$ Corpus\footnote{\url{wit3.fbk.eu}} \cite{cettolo12}, a collection of transcripts of TED talks.  We use 150k sentence pairs for training, 5k for development and 50k for testing. 

\textbf{Pre-processing.} Before assigning sense labels, we tokenize all the texts and identify the parts of speech using the Stanford POS tagger\footnote{\url{nlp.stanford.edu/software} \label{footnote_1}}. Then, we filter out the stopwords and the nouns which are proper names according to the Stanford Name Entity Recognizer$^{\ref{footnote_1}}$. Furthermore, we convert the plural forms of nouns to their singular forms and the verb forms to infinitives using the stemmer and lemmatizer from NLTK\footnote{\url{www.nltk.org}}, which is essential because WordNet has description entries only for base forms. The pre-processed text is used for assigning sense labels to each occurrence of a noun or verb that has more than one sense in WordNet. 

\textbf{K-means settings.} Unless otherwise stated, we adopt the following settings in the $k$-means algorithm, with the implementation provided in Scikit-learn \cite{Pedregosa11}.  We use the definition of each sense for initializing the centroids, and later compare this choice with the use of examples.  We set $k_t$, the initial number of clusters, to the number of WordNet senses of each ambiguous word type $W_t$, and set the window size for the context surrounding each occurrence to $c=8$.

\textbf{Neural MT.} We build upon the attention-based neural translation model \cite{bahdanau2014neural} from the OpenNMT toolkit \cite{2017opennmt}\footnote{\url{www.opennmt.net}}. We use long short-term memory units (LSTM) and not gated recurrent units (GRU). For the proposed \emph{ATT} and \emph{ATT}$_\mathit{ini}$ models, we add an external attention layer before the encoder, but do not otherwise alter the internals of the NMT model.

We set the source and target vocabulary sizes to 50,000 and the dimension of word embeddings to 500, which is recommended for OpenNMT, so as to reach a strong baseline.  For the \emph{ATT}$_\mathit{ini}$ model, since the embeddings from word2vec used for initialization have only 300 dimensions, we randomly pick up a vector with 200 dimensions within range [-0.1,0.1] and concatenate it with the vector from word2vec to reach the required number of dimensions, ensuring a fair comparison.

It takes around 15 epochs (25-30 hours on Idiap's GPU cluster) to train each of the five NMT models: the baseline and our four proposals.  The \emph{AVG} model takes more time for training (around 40 hours) since we use additional weights and senses for each token.  In fact, we limit the number of senses for \emph{AVG} to 5 per word type, after observing that in WordNet there are fewer than 100 words with more than 5 senses.

\begin{table*}[ht] \small
\centering
\begin{tabular}{|c|c|c|c|c|c|c|c|c|c|}
\hline
\multirow{2}{*}{Pair}  & \multirow{2}{*}{Initialization} & \multicolumn{5}{c|}{BLEU}                                                  & \multicolumn{3}{c|}{$\rho$ (\%)}                \\ \cline{3-10} 
                       &                           & Baseline               & Graph                  & CRP   & $k$-means & Oracle & Graph                  & CRP   & $k$-means \\ \hline
\multirow{2}{*}{EN/ZH} & Definitions               & \multirow{2}{*}{15.23} & \multirow{2}{*}{15.31} & 15.31 & \textbf{15.54}   & 16.24  & \multirow{2}{*}{+0.20} & +0.27 & \textbf{+2.25}  \\  
                       & Examples                  &                        &                        & 15.28 & 15.41   & 15.85  &                        & +0.13 & +1.60  \\ \hline
\multirow{2}{*}{EN/DE} & Definitions               & \multirow{2}{*}{19.72} & \multirow{2}{*}{19.74} & 19.69 & \textbf{20.23}   & 20.99  & \multirow{2}{*}{-0.07} & -0.19 & \textbf{+3.96}  \\  
                       & Examples                  &                        &                        & 19.74 & 19.87   & 20.45  &                        & -0.12 & +2.15  \\ \hline
\end{tabular}
\caption{Performance of the WSD+SMT factored system for two language pairs from WIT$\mathrm{3}$, with three clustering methods and two initializations.}
\label{tab:bleu-wit}
\end{table*}

\begin{table*}[ht]\small
\centering
\begin{tabular}{|l|c|c|c|c|c|c|c|c|c|c|}
\hline
\multirow{2}{*}{System} & \multicolumn{3}{c|}{V-score} & \multicolumn{3}{c|}{$F_{1}$-score} & \multicolumn{3}{c|}{Average} &  \multirow{2}{*}{C} \\ \cline{2-10}
                      & All      & Nouns     & Verbs     & All      & Nouns    & Verbs & All & Nouns & Verbs   &  \\ \hline
UoY                   & 15.70    & 20.60     & 8.50      & 49.80     & 38.20    & 66.60   & 32.75   & 29.40   & \bf{37.50}    & 11.54  \\
KCDC-GD               &  6.90    & 5.90      & 8.50      & 59.20     & 51.60    & 70.00   & 33.05   & 28.70   & 39.20    &  2.78  \\ 
Duluth-Mix-Gap        & 3.00     & 2.90      & 3.00      & 59.10     & 54.50    & 65.80   & 31.05   & 29.70   & 34.40    &  1.61  \\ \hline
$k$-means+definitions& 13.65   & 14.70     & 12.60     &  56.70    & 53.70    & 59.60   & \bf{35.20}  & \bf{34.20}  & 36.10 & 4.45\\ 
$k$-means+examples  & 11.35    & 11.00     & 11.70     & 53.25     & 47.70    & 58.80   & 32.28   & 29.30   &  35.25   &  3.58\\ 
CRP + definitions     & 1.45    & 1.50      &  1.45     & 64.80     & 56.80    & 72.80   & 33.13   & 29.15   &  37.10   &  1.80\\ 
CRP + examples        & 1.20     & 1.30      &  1.10     & 64.75     & 56.80    & 72.70   & 32.98   & 29.05   &  36.90   &  1.66\\
Graph + definitions     & 11.30    & 11.90      &  10.70     & 55.10     & 52.80    & 57.40   & 33.20   & 32.35   &  34.05   &  2.63\\ 
Graph + examples        & 9.05     & 8.70      &  9.40     & 50.15     & 45.20    & 55.10   & 29.60   & 26.96   &  32.25   &  2.08\\ \hline
\end{tabular}
\caption{WSD results from three SemEval 2010 systems and our six systems, in terms of $V$-score, $F_{1}$ score and their average. `C' is the average number of clusters. The adaptive $k$-means using definitions outperforms the others on the average of $V$ and $F_1$, when considering both nouns and verbs, or nouns only.  The SemEval systems are UoY \protect\cite{korkontzelos2010uoy}; KCDC-GD \protect\cite{kern2010kcdc}; and Duluth-Mix-Gap \protect\cite{pedersen2010duluth}.}
\label{table:wsd_system_comparison} 
\vspace{-3mm}
\end{table*}


\textbf{Evaluation metrics.}   
For the evaluation of intrinsic WSD performance, we use the $V$-score, the $F_{1}$-score, and their average, as used for instance at SemEval 2010 \cite{manandhar2010semeval}. The $V$-score is the weighted harmonic mean of homogeneity and completeness (favoring systems generating more clusters than the reference), while the $F_{1}$-score measures the classification performance (favoring systems generating fewer clusters).  Therefore, the ranking metric for SemEval 2010 is the average of the two.

We select the optimal model configuration based on MT performance on development sets, as measured with the traditional \textit{multi-bleu} score \cite{papineni2002bleu}.  Moreover, to estimate the impact of WSD on MT, we also measure the actual impact on the nouns and verbs that have several WordNet senses, by counting how many of them are translated exactly as in the reference translation.  To quantify the difference with the baseline, we use the following coefficient.  First, for a certain set of tokens in the source data, we note as $N_\mathrm{improved}$ the number of tokens which are translated by our system with the same token as in the reference translation, but are translated differently by the baseline system.  Conversely, we note as $N_\mathrm{degraded}$ the number of tokens which are translated by the baseline system as in the reference, but differently by our system\footnote{The values of $N_\mathrm{improved}$ and $N_\mathrm{degraded}$ are obtained using automatic word alignment.  They do not capture, of course, the intrinsic correctness of a candidate translation, but only its identity or not with one reference translation.}.  We use the normalized coefficient $\rho = (N_\mathrm{improved} - N_\mathrm{degraded})/T$, where $T$ is the total number of tokens, as a metric to specifically evaluate the translation of words submitted to WSD.

For all tables we mark in bold the best score per condition. For MT scores in Tables~\ref{tab:bleu_all_result}, \ref{tab:bleu_wmt}, and~\ref{tab:bleu_de_compare}, we show the improvement over the baseline and its significance based on two confidence levels: either $p <$ 0.05 (indicated with a `$\dagger$') or  $p <$ 0.01 (`$\ddagger$').  $P$-values larger than 0.05 are treated as not significant and are left unmarked.

\section{Optimal Values of the Parameters}
\label{sec:optimal-parameters}

\subsection{Best WSD Method Based on BLEU}

We first select the optimal clustering method and its initialization settings, in a series of experiments with statistical MT over the WIT$^\mathrm{3}$ corpus, extending and confirming our previous results \citep{pu-pappas-popescubelis:2017:WMT}.  In Table~\ref{tab:bleu-wit}, we present the BLEU and $\rho$ scores of our previous WSD+SMT system for the three clustering methods, initialized with vectors either from the WordNet definitions or from examples, for two language pairs.  We also provide BLEU scores of baseline systems and of oracle ones, i.e.\ using correct senses as factors.  The best method is $k$-means and the best initialization is with the vectors of definitions.  All values of $\rho$ show improvements over the baseline, with up to 4\% for $k$-means on DE/EN.

Moreover, we found that random initializations under-perform with respect to definitions or examples.  For a fair comparison, we set the number of clusters equal either to the number of synsets with definitions or with examples, for each word type, and obtained BLEU scores on EN/ZH of 15.34 and 15.27 respectively -- hence lower than 15.54 and 15.41 in Table~\ref{tab:bleu-wit}.  We investigated earlier \citep{pu-pappas-popescubelis:2017:WMT} the effect of the context window surrounding each ambiguous token, and found with the WSD+SMT factored system on EN/ZH WIT$^\mathrm{3}$ data that the optimal size was~8, which we use here as well.

\subsection{Best WSD Method Based on V/F1 Scores}

Table~\ref{table:wsd_system_comparison} shows our WSD results in terms of $V$-score and $F_{1}$-score, comparing our methods (six lines at the bottom) with other significant systems that participated in the SemEval 2010 shared task \cite{manandhar2010semeval}.\footnote{We provide comparisons with more systems from SemEval in our previous paper \citep{pu-pappas-popescubelis:2017:WMT}.}  The adaptive $k$-means initialized with definitions has the highest average score (35.20) and ranks among the top systems for most of the metrics individually.  Moreover, the adaptive $k$-means method finds on average 4.5 senses per word type, which is very close to the ground-truth value of~4.46.  Overall, we observed that $k$-means infers fewer senses per word type than WordNet.  These results show that $k$-means WSD is effective and provides competitive performance against other weakly supervised alternatives (CRP or Random Walk) and even against SemEval WSD methods, but using additional knowledge not available to SemEval participants.

\begin{table}[ht] \small
\centering
\begin{tabular}{|l|l|}
\hline
System and settings & BLEU  \\ \hline
Baseline                                                 & 29.55 \\ 
\emph{TOP}                                               & 29.63 \textit{(+0.08)}\\  
\emph{AVG} with linear norm.\ in Eq.~\ref{eq_logistic}   & 29.67 \textit{(+0.12)}\\  
\emph{AVG} with logistic norm.\ in Eq.~\ref{eq_logistic} & \textbf{30.15} \textit{(+0.60)}\\ 
\emph{ATT} with \textsc{null} label                      & 29.80 \textit{(+0.33)}\\ 
\emph{ATT} with word used as label                       & \textbf{30.23} \textit{(+0.68)}\\  
\emph{ATT}$_\mathit{ini}$ with $\mathbf{u}_{i}^{T}W\bm{\mu}_{ij}$ in Eq.~\ref{eq:attention} & 29.94 \textit{(+0.39)} \\ 
\emph{ATT}$_\mathit{ini}$ with $tanh$ in Eq.~\ref{eq:attention} & \textbf{30.61} \textit{(+1.06)} \\ \hline 
\end{tabular}
\caption{Performance of various WSD+NMT configurations on a EN/FR subset of Europarl, with variations wrt.\ baseline.  We select the settings with the best performance (bold) for our final experiments in Section~\ref{sec:mt-results}.}
\label{tab:nmt_settings}
\vspace{-3mm}
\end{table}

\subsection{Selection of WSD+NMT Model}

To compare several options of the WSD+NMT systems, we trained and tested them on a subset of EN/FR Europarl (a smaller dataset shortened the training times).  The results are shown in Table~\ref{tab:nmt_settings}.  For the \emph{AVG} model, the logistic normalization in Eq.~\ref{eq_logistic} works better than the linear one.  For the \emph{ATT} model, we compared two different labeling approaches for tokens which do not have multiple senses: either use the same \textsc{null} label for all tokens, or use the word itself as a label for its sense; the second option appeared to be the best.
Finally, for the \emph{ATT}$_\mathit{ini}$ model, we compared the two options for the attention function in Eq.~\ref{eq:attention}, and found that the formula with $tanh$ is the best.  In what follows, we use these settings for the \emph{AVG} and \emph{ATT} systems.

\begin{table*}[ht]\small
\centering
\begin{tabular}{|l|c|c|c|c|c|} 
\hline
& EN/FR                  & EN/DE                  & EN/ZH                  & EN/ES                  & EN/NL \\ \hline
SMT baseline & 31.96                  & 20.78                  & 23.25                  & 39.95                  & 23.56 \\ 
Graph    & 32.01 \textit{(+.05)} & 21.17 \textit{(+.39)} & 23.47 \textit{(+.22)} & 40.15 \textit{(+.20)} & 23.74 \textit{(+.18)} \\  
CRP      & 32.08 \textit{(+.12)} & 21.19 \textit{(+.41)} $\dagger$ & 23.55 \textit{(+.29)} & 40.14 \textit{(+.19)} & 23.79 \textit{(+.23)} \\ 
$k$-means& 32.20 \textit{(+.24)} & 21.32 \textit{(+.54)} $\dagger$ & 23.69 \textit{(+.44)} $\dagger$ & 40.37 \textit{(+.42)} $\dagger$ & 23.84 \textit{(+.26)}  \\ \hline

NMT baseline  & 34.60                 & 25.80                 & 27.07                 & 44.09                 & 24.79 \\ 
$k$-means + \emph{TOP}& 34.52 \textit{($-$.08)} & 25.84 \textit{(+.04)} & 26.93 \textit{($-$.14)} & 44.14 \textit{(+.05)} & 24.71 \textit{($-$.08)} \\ 
$k$-means + \emph{AVG}& 35.17 \textit{(+.57)} $\dagger$ & 26.47 \textit{(+.67)} $\dagger$ & 27.44 \textit{(+.37)} & 45.05 \textit{(+.97)} $\ddagger$ & 25.04 \textit{(+.25)} \\ 
None + \emph{ATT}& 35.32 \textit{(+.72)} $\ddagger$ & 26.50 \textit{(+.70)} $\ddagger$ & 27.56 \textit{(+.49)} $\dagger$ & 44.93 \textit{(+.84)} $\ddagger$ & 25.36 \textit{(+.57)} $\dagger$ \\ 
$k$-means + \emph{ATT}$_\mathit{ini}$&\textbf{35.78 \textit{(+1.18)}} $\ddagger$ & \textbf{26.74 \textit{(+.94)}} $\ddagger$ & \textbf{27.84 \textit{(+.77)}} $\ddagger$ & \textbf{45.18 \textit{(+1.09)}} $\ddagger$ & \textbf{25.65 \textit{(+.86)}} $\ddagger$ \\ \hline 
\end{tabular}
\caption{BLEU scores of our sense-aware NMT systems over five language pairs: \emph{ATT}$_\mathit{ini}$ is the best one among SMT and NMT systems.  Significance testing is indicated by $\dagger$ for $p <$ 0.05 and $\ddagger$ for $p <$ 0.01.}
\label{tab:bleu_all_result}
\vspace{-3mm}
\end{table*}

\section{Results}
\label{sec:mt-results}

We first evaluate our sense-aware models with smaller data sets (ca.\ 500k lines) for five language pairs with English as source. We evaluate them through both automatic measures and human assessment. Later on, we evaluate our sense-aware NMT models with larger WMT data sets to enable a better comparison with other related approaches.

\textbf{BLEU scores.}  
Table~\ref{tab:bleu_all_result} displays the performance of both sense-aware phrase-based and neural MT systems with the training sets of 500k lines listed in Table~\ref{mt_data} on five language pairs. Specifically, we compare several approaches which integrate word sense information in SMT and NMT. The best hyper-parameters are those found above, for each of the WSD+NMT combination strategies, in particular the $k$-means method for WSD+SMT, and the \emph{ATT}$_\mathit{ini}$ method for WSD+NMT, i.e.\ the attention-based model of senses initialized with the output of $k$-means clustering.

\textbf{Comparisons with baselines.}
Table~\ref{tab:bleu_all_result} shows that our WSD+NMT systems perform consistently better than the baselines, with the largest improvements achieved by NMT on EN/FR and EN/ES.  The neural systems outperform the phrase-based statistical ones \citep{pu-pappas-popescubelis:2017:WMT}, which are shown for comparison in the upper part of the table.  

We compare our proposal to the recent system proposed by \citet{yang2017multi}, on the 500k-line EN/FR Europarl dataset (the differences between their system and ours are listed in Section~\ref{sec:related-work}).  We carefully implemented their model by following their paper, since their code is not available.  Using the sense embeddings of the multi-sense skip-gram model (MSSG) \citep{arvind2014efficient} as they do, and training for 6 epochs as in their study, our implementation of their model reaches only 31.05 BLEU points.  When increasing the training stage until convergence (15 epochs), the best BLEU score is 34.52, which is still below our NMT baseline of 34.60.  We also found that the initialization of embeddings with MSSG brings less than 1 BLEU point improvement with respect to random initializations (which scored 30.11 over 6 epochs and 33.77 until convergence), while Yang et al.\ found a 1.3--2.7 increase on two different test sets. In order to better understand the difference, we tried several combinations of their model with ours.  We obtain a BLEU score of 35.02 by replacing their MSSG sense specification model with our adaptive $k$-means approach, and a BLEU score of 35.18 by replacing our context calculation method (averaging word embeddings within one sentence) with their context vector generation method, which is computed from the output of a bi-directional RNN.  In the end, the best BLEU score on this EN/FR data set (35.78 as shown in Table~\ref{tab:bleu_all_result}, column~1, last line) is reached by our system with its best options.

\textbf{Lexical choice.}
Using word alignment, we assess the improvement brought by our systems with respect to the baseline in terms of the number of words -- here, WSD-labeled nouns and verbs -- that are translated exactly as in the reference translation (modulo alignment errors).  These numbers can be arranged in a confusion matrix with four values: the words translated correctly (i.e., as in the reference) by both systems, those translated correctly by one system but incorrectly by the other one, and vice-versa, and those translated incorrectly by both.

Table~\ref{tab:correct_incorrect} shows the confusion matrix for our sense-aware NMT with the \emph{ATT}$_\mathit{ini}$ model \emph{versus} the NMT baseline over the Europarl test data.  The net improvement, i.e.\ the fraction of words improved by our system minus those degraded,\footnote{Explicitly, improvements are (system-correct \& baseline-incorrect) minus (system-incorrect \& baseline-correct), and degradations the converse difference.} appears to be +2.5\% for EN/FR and +3.6\% for EN/ES. For comparison, we show the results of the WSD+SMT system \emph{versus} the SMT baseline in the lower part of Table~\ref{tab:correct_incorrect}: the improvement is smaller, at +1.4\% for EN/FR and +1.5\% for EN/ES.   Therefore, the \emph{ATT}$_\mathit{ini}$ NMT model brings higher benefits over the NMT baseline than the WSD+SMT factored model, although the NMT baseline is stronger than the SMT one (see Table~\ref{tab:bleu_all_result}).

\begin{table}[t]\small
\setlength{\tabcolsep}{3pt}
\centering
\begin{tabular}{|l|l|cc|cc|}
\hline
\multicolumn{2}{|l|}{\multirow{3}{*}{}} & \multicolumn{4}{c|}{Baselines}                                                       \\ \cline{3-6} 
\multicolumn{2}{|l|}{}                  & \multicolumn{2}{c|}{EN/FR}               & \multicolumn{2}{c|}{EN/ES}               \\ \cline{3-6} 
\multicolumn{2}{|l|}{}                  & \multicolumn{1}{c|}{Correct} & Incorrect & \multicolumn{1}{c|}{Correct} & Incorrect \\ \hline
\multicolumn{1}{|l|}{WSD+}      & C.      & 134,552                      & 17,145    & 146,806                      & 16,523    \\
\multicolumn{1}{|l|}{NMT}      & I.      & 10,551                       & 101,228   & 8,183                        & 58,387    \\ \hline
\hline
\multicolumn{1}{|l|}{WSD+}      & C.      & \multicolumn{1}{c}{124,759} & 13,408    & 139,800                      & 11,194    \\
\multicolumn{1}{|l|}{SMT}               & I.      & \multicolumn{1}{c}{9,676}   & 115,633   & 7,559                        & 71,346    \\ \hline
\end{tabular}
\caption{Confusion matrix for our WSD+NMT (\emph{ATT}$_\mathit{ini}$) system and our WSD+SMT system against their respective baselines (NMT and SMT), over the Europarl test data, for two language pairs.} 
\label{tab:correct_incorrect}
\end{table}

\textbf{Human assessment.}
To compare our systems against baselines we also consider a human evaluation of the translation of words with multiple senses (nouns or verbs).  The goal is to capture more precisely the correct translations that are, however, different from the reference. 

Given the cost of the procedure, one evaluator with good knowledge of EN and FR rated the translations of four word types that appear frequently in the test set and have multiple possible senses and translations into French.  These words are: `deal' (101 tokens), `face' (84), `mark' (20), and `subject' (58).  Two translations of `deal' are exemplified in Section~\ref{intro}.  

For each occurrence, the evaluator sees the source sentence, the reference translation, and the outputs of the NMT baseline and the \emph{ATT}$_\mathit{ini}$ in random order, so that the system cannot be identified.  The two translations of the considered word are rated as good, acceptable, or wrong. We submit only cases in which the two translations differ, to minimize the annotation effort with no impact on the comparison between systems.

\begin{figure}[t]  
\centering
\includegraphics[width=0.21\textwidth,trim={0.71cm 0.74cm 4cm 0},clip]{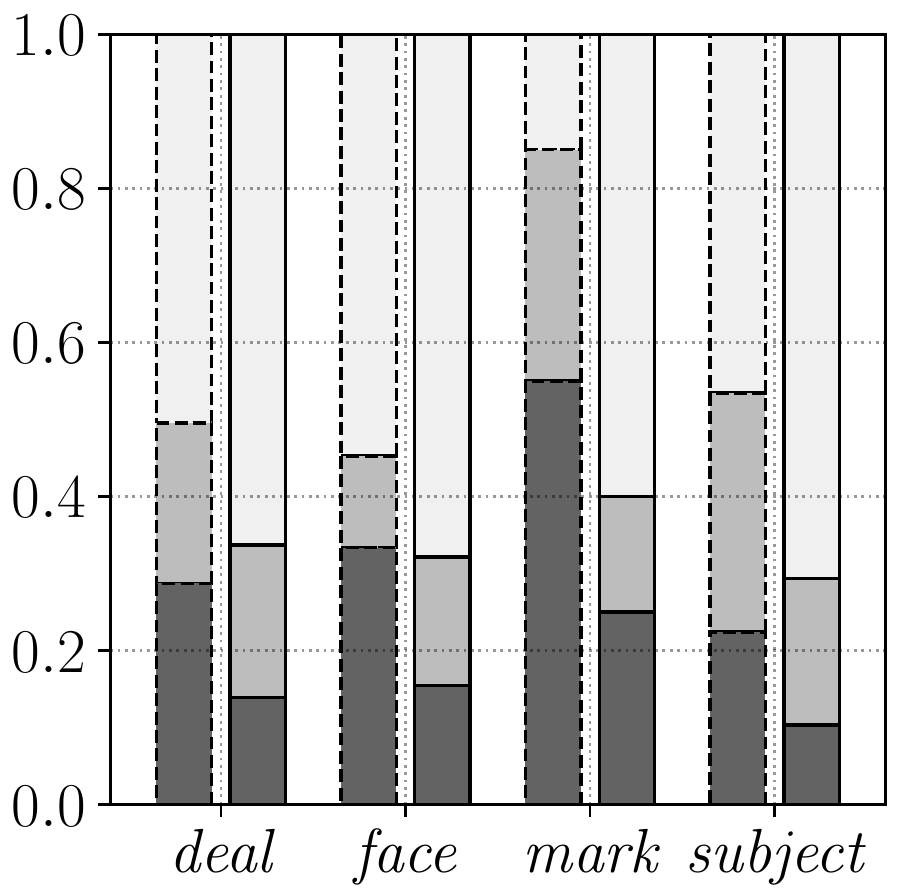}~~~~~\includegraphics[width=0.21\textwidth,trim={0.62cm 0.7cm 2.8cm 0},clip]{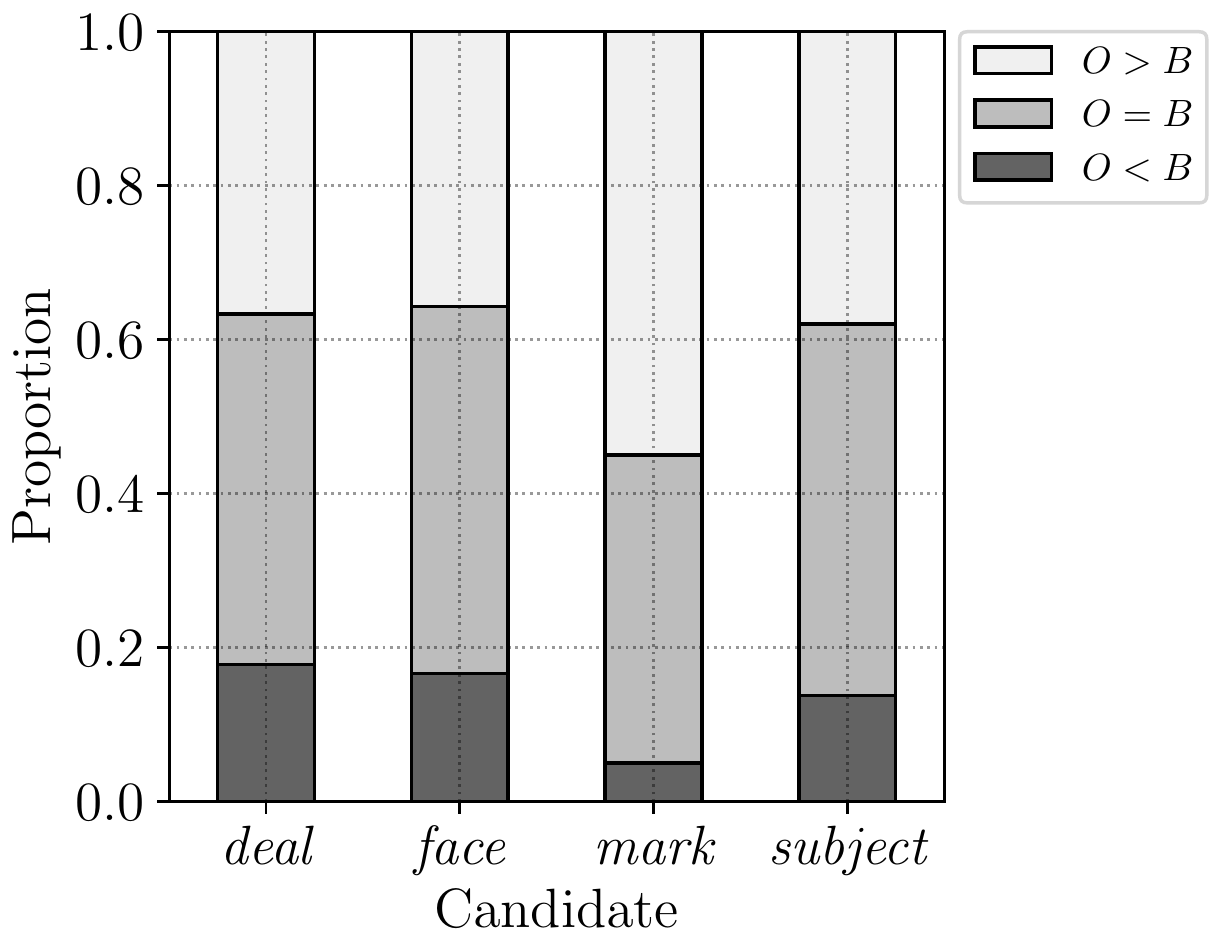}\\
\vspace{.2em}
{\small ~~~~(a)~System ratings.~~~~~~~~(b)~Comparative scores.}\vspace{-.5em}
\caption{Human comparison of the EN/FR translations of four word types. (a)~Proportion of good (light gray), acceptable (middle gray) and wrong (dark gray) translations per word and system (baseline left, \emph{ATT}$_\mathit{ini}$ right, for each word). (b)~Proportion of translations in which \emph{ATT}$_\mathrm{ini}$ is better (light gray), equal (middle gray) or worse (dark gray) than the baseline.}
\label{fig:manual_evaluation}
\vspace{-3mm}
\end{figure}

Firstly, Figure~\ref{fig:manual_evaluation}~(a) shows that \emph{ATT}$_\mathit{ini}$ has a higher proportion of good translations, and a lower proportion of wrong ones, for all four words.  The largest difference is for `subject', where \emph{ATT}$_\mathit{ini}$ has 75\% good translations and the baseline only 46\%; moreover, the baseline has 22\% errors and \emph{ATT}$_\mathit{ini}$ has only 9\%.  
Secondly, Figure~\ref{fig:manual_evaluation}~(b) shows the proportions of tokens, for each type, for which \emph{ATT}$_\mathit{ini}$ was respectively better, equal, or worse than the baseline.  Again, for each of the four words, there are far more improvements brought by \emph{ATT}$_\mathit{ini}$ than degradations.  On average, 40\% of the occurrences are improved and only 10\% are degraded.  

\textbf{Results on WMT datasets}. 
To demonstrate that our findings generalize to larger datasets, we report results on three datasets provided by the WMT conference (see Section~\ref{sec:data}), namely for EN/DE, EN/ES and EN/FR. Tables~\ref{tab:bleu_wmt} and~\ref{tab:bleu_de_compare} show the results of our proposed NMT models on these test sets.  The results in Table~\ref{tab:bleu_wmt} confirm that our sense-aware NMT models improve significantly the translation quality also on larger datasets, which permit stronger baselines.  Comparing these results to the ones from Table~\ref{tab:bleu_all_result}, we even conclude that our models trained on larger, mixed-domain datasets achieve higher improvements than the models trained on smaller, domain-specific datasets (Europarl). This clearly shows that our sense-aware NMT models are beneficial on both narrow and broad domains. 

\begin{table*}[ht]\small
\centering
\begin{tabular}{|l|cc|cc|}
\hline
\multirow{2}{*}{} & \multicolumn{2}{c|}{EN/FR}  & \multicolumn{2}{c|}{EN/ES} \\ \cline{2-5}  
                  & NT12         & NT13        & NT12         & NT13        \\ \hline
Baseline          & 29.09        & 29.60         & 32.66        & 29.57       \\ 
None + \emph{ATT}              & 29.47 \textit{(+.38)}        & 30.21 \textit{(+.61)} $\dagger$     & 33.15 \textit{(+.49)} $\dagger$      & 30.27 \textit{(+.70)} $\ddagger$     \\ 
$k$-means + \emph{ATT}$_\mathit{ini}$         & \textbf{30.26} \textit{(+1.17)} $\ddagger$      & \textbf{30.95} \textit{(+.1.35)} $\ddagger$   & \textbf{34.14}  \textit{(+1.48)} $\ddagger$     & \textbf{30.67} \textit{(+1.1)} $\ddagger$     \\ 
\hline
\end{tabular}
\caption{BLEU scores on WMT NewsTest 2012 and 2013 (NT) test sets for two language pairs. Significance testing is indicated by $\dagger$ for $p <$ 0.05 and $\ddagger$ for $p <$ 0.01.}
\label{tab:bleu_wmt}
\end{table*}

Finally, we compare our model to several recent NMT models which make use of contextual information, thus sharing a similar overall goal to our study.  Indeed, the model proposed by \citet{choi2017context} attempts to improve NMT by integrating context vectors associated to source words into the generation process during decoding. The model proposed by \citet{zhang2017context} is aware of previous attended words on the source side in order to better predict which words will be attended in future.  The self-attentive residual decoder designed by \citet{werlen2018self} leverages the contextual information from previously translated words on the target side. BLEU scores on the English-German pair shown in Table~\ref{tab:bleu_de_compare} demonstrate that our baseline is strong and that our model is competitive with respect to recent models that leverage contextual information in different ways.

\begin{table*}[ht]\small
\setlength{\tabcolsep}{3pt}
\centering
\begin{tabular}{|l|c|c|}
\hline
NMT model                                & NT14   & NT15  \\ \hline
Context-dependent \cite{choi2017context} & -      & 21.99 \\ 
Context-aware \cite{zhang2017context}    & 22.57  & -     \\ 
Self-attentive \cite{werlen2018self}     & 23.2   & 25.5  \\ \hline
Baseline                                 & 22.79  & 24.94  \\ 
None + \emph{ATT}                        & 23.34 $\dagger$  & 25.28 \\ 
$k$-means + \emph{ATT}$_\mathit{ini}$ & \textbf{23.85} \textit{(+1.14)} $\ddagger$ & \textbf{25.71} \textit{(+0.77)} $\ddagger$ \\ \hline
\end{tabular}
\caption{BLEU score on English-to-German translation over the WMT NewsTest (NT) 2014 and 2015 test sets. Significance testing is indicated by $\dagger$ for $p <$ 0.05 and $\ddagger$ for $p <$ 0.01. The highest score per column is in bold.}
\label{tab:bleu_de_compare}
\end{table*}

\section{Related Work}
\label{sec:related-work}

Word sense disambiguation aims to identify the sense of a word appearing in a given context \cite{agirre2007word}. Resolving word sense ambiguities should be useful, in particular, for lexical choice in MT. An initial investigation found that a statistical MT system which makes use of off-the-shelf WSD does not yield significantly better quality translations than a SMT system not using it \cite{carpuat2005word}.  However, several studies \cite{Vickrey2005,cabezas2005using,chan2007word,carpuat2007improving} reformulated the task of WSD for SMT and showed that integrating the ambiguity information generated from modified WSD improved SMT by 0.15--0.57 BLEU points compared to baselines.

Recently, \citet{tang2016improving} used only the super-senses from WordNet (coarse-grained semantic labels) for automatic WSD, using maximum entropy classification or sense embeddings learned using word2vec.  When combining WSD with SMT using a factored model, Tang et al.\ improved BLEU scores by 0.7 points on average, though with large differences between their three test subsets (IT Q\&A pairs).

Although the above-mentioned reformulations of the WSD task proved helpful for SMT, they did not determine whether actual source-side senses are helpful or not for end-to-end SMT.  \citet{xiong2014sense} attempted to answer this question by performing self-learned word sense induction instead of using pre-specified word senses as traditional WSD does. However, they created the risk of discovering sense clusters which do not correspond to the senses of words actually needed for MT. Hence, they left open an important question, namely whether WSD based on semantic resources such as WordNet \citep{wordnet1998book} can be successfully integrated with SMT.

Several studies integrated sense information as features to SMT, either obtained from the sense graph provided by WordNet \cite{neale2016word} or generated from both sides of word dependencies \cite{su2015graph}. However, apart from the sense graph, WordNet provides also textual information such as sense definitions and examples, which should be useful for WSD, but were not used in the above studies.  In previous work \citep{pu-pappas-popescubelis:2017:WMT}, we used this information to perform sense induction on source-side data using $k$-means and demonstrated improvement with factored phrase-based SMT but not NMT.

Neural MT became the state of the art \citep{sutskever2014sequence,bahdanau2014neural}. Instead of working directly at the discrete symbol level as SMT, it projects and manipulates the source sequence of discrete symbols in a continuous vector space. However, NMT generates only one embedding for each word type, regardless of its possibly different senses, as analyzed e.g.\ by \citet{Hill2017}.  
Several studies proposed efficient non-parametric models for monolingual word sense representation \cite{arvind2014efficient,li-jurafsky:2015:EMNLP,bartunov2016breaking,liu2017handling}, but left open the question whether sense representations can help neural MT by reducing word ambiguity.  Recent studies integrate the additional sense assignment with neural MT based on these approaches, either by adding such sense assignments as additional features \citep{lauraWSDforNMT} or by merging the context information on both sides of parallel data for encoding and decoding \citep{choi2017context}.

\citet{yang2017multi} recently proposed to add sense information by using weighted sense embeddings as input to neural MT.
The sense labels were generated by a multi-sense skip-gram model (MSSG) \cite{arvind2014efficient}, and the context vector used for sense weight generation  was computed from the output of a bi-directional RNN. Finally, the weighted average sense embeddings were used in place of the word embedding for the NMT encoder.  The numerical results given in Section~\ref{sec:mt-results} above show that our options for using sense embeddings  outperform Yang et al.'s proposal.  In fact, their approach even performed below the NMT baseline on our EN/FR dataset.  We conclude that adaptive $k$-means clustering is better than MSSG for use in NMT, and that concatenating the word embedding and its sense vector as input for the RNN encoder is better than just using the sense embedding for each token.  In terms of efficiency, \citet{yang2017multi} need an additional bi-directional RNN to generate the context vector for each input token, while we compute the context vector by averaging the embeddings of the neighboring tokens.  This slows down the training of the encoder by a factor of 3, which may explain why they only trained their model for 6 epochs.

\section{Conclusion}
We presented a neural MT system enhanced with an attention-based method to represent multiple word senses, making use of a larger context to disambiguate words that have various possible translations.  We proposed several adaptive context-dependent clustering algorithms for WSD and combined them in several ways with NMT -- following our earlier experiments with SMT \citep{pu-pappas-popescubelis:2017:WMT} -- and found that they had competitive WSD performance on data from the SemEval 2010 shared task.  

For NMT, the best performing method used the output of $k$-means to initialize the sense embeddings that are learned by our system.  In particular, it appeared that learning sense embeddings for NMT is better than using embeddings learned separately by other methods, although such embeddings may be useful for initialization. Our experiments with five language pairs showed that our sense-aware NMT systems consistently improve over strong NMT baselines, and that they specifically improve the translation of words with multiple senses.

In the future, our approach to sense-aware NMT could be extended to other NMT architectures such as the Transformer network proposed by \citet{vaswani2017attention}.  As was the case with the LSTM-based architecture studied here, the Transformer network does not explicitly model or utilize the sense information of words, and, therefore, we hypothesize that its  performance could also be improved by using our sense integration approaches. To encourage further research in sense-aware NMT, our code is made available at \url{https://github.com/idiap/sense_aware_NMT}.

\section*{Acknowledgments}
The authors are grateful for support from the Swiss National Science Foundation through the MODERN Sinergia project on Modeling Discourse Entities and Relations for Coherent Machine Translation, grant n.\ 147653 (\url{www.idiap.ch/project/modern}), and from the European Union through the SUMMA Horizon 2020 project on Scalable Understanding of Multilingual Media, grant n.\ 688139 (\url{www.summa-project.eu}).  The authors would like to thank the TACL editors and reviewers for their helpful comments and suggestions.

\bibliographystyle{acl_natbib}
\bibliography{mybib}

\end{document}